\documentclass[a4paper,conference]{IEEEtran}
\IEEEoverridecommandlockouts
\usepackage{cite}
\usepackage{amsmath,amssymb,amsfonts}
\usepackage{algorithmic}
\usepackage{graphicx}
\usepackage{textcomp}
\usepackage{xcolor}
\usepackage{tikz} 
\usepackage{pgfplots}
\usetikzlibrary{spy,calc}
\usepackage{subcaption}

\newif\ifblackandwhitecycle
\gdef\patternnumber{0}

\pgfkeys{/tikz/.cd,
    zoombox paths/.style={
        draw=orange,
        very thick
    },
    black and white/.is choice,
    black and white/.default=static,
    black and white/static/.style={ 
        draw=white,   
        zoombox paths/.append style={
            draw=white,
            postaction={
                draw=black,
                loosely dashed
            }
        }
    },
    black and white/static/.code={
        \gdef\patternnumber{1}
    },
    black and white/cycle/.code={
        \blackandwhitecycletrue
        \gdef\patternnumber{1}
    },
    black and white pattern/.is choice,
    black and white pattern/0/.style={},
    black and white pattern/1/.style={    
            draw=white,
            postaction={
                draw=black,
                dash pattern=on 2pt off 2pt
            }
    },
    black and white pattern/2/.style={    
            draw=white,
            postaction={
                draw=black,
                dash pattern=on 4pt off 4pt
            }
    },
    black and white pattern/3/.style={    
            draw=white,
            postaction={
                draw=black,
                dash pattern=on 4pt off 4pt on 1pt off 4pt
            }
    },
    black and white pattern/4/.style={    
            draw=white,
            postaction={
                draw=black,
                dash pattern=on 4pt off 2pt on 2 pt off 2pt on 2 pt off 2pt
            }
    },
    zoomboxarray inner gap/.initial=5pt,
    zoomboxarray columns/.initial=2,
    zoomboxarray rows/.initial=2,
    subfigurename/.initial={},
    figurename/.initial={zoombox},
    zoomboxarray/.style={
        execute at begin picture={
            \begin{scope}[
                spy using outlines={%
                    zoombox paths,
                    width=\imagewidth / \pgfkeysvalueof{/tikz/zoomboxarray columns} - (\pgfkeysvalueof{/tikz/zoomboxarray columns} - 1) / \pgfkeysvalueof{/tikz/zoomboxarray columns} * \pgfkeysvalueof{/tikz/zoomboxarray inner gap} -\pgflinewidth,
                    height=\imageheight / \pgfkeysvalueof{/tikz/zoomboxarray rows} - (\pgfkeysvalueof{/tikz/zoomboxarray rows} - 1) / \pgfkeysvalueof{/tikz/zoomboxarray rows} * \pgfkeysvalueof{/tikz/zoomboxarray inner gap}-\pgflinewidth,
                    magnification=3,
                    every spy on node/.style={
                        zoombox paths
                    },
                    every spy in node/.style={
                        zoombox paths
                    }
                }
            ]
        },
        execute at end picture={
            \end{scope}
            \node at (image.north) [anchor=north,inner sep=0pt] {\subcaptionbox{\label{\pgfkeysvalueof{/tikz/figurename}-image}}{\phantomimage}};
            \node at (zoomboxes container.north) [anchor=north,inner sep=0pt] {\subcaptionbox{\label{\pgfkeysvalueof{/tikz/figurename}-zoom}}{\phantomimage}};
     \gdef\patternnumber{0}
        },
        spymargin/.initial=0.5em,
        zoomboxes xshift/.initial=1,
        zoomboxes right/.code=\pgfkeys{/tikz/zoomboxes xshift=1},
        zoomboxes left/.code=\pgfkeys{/tikz/zoomboxes xshift=-1},
        zoomboxes yshift/.initial=0,
        zoomboxes above/.code={
            \pgfkeys{/tikz/zoomboxes yshift=1},
            \pgfkeys{/tikz/zoomboxes xshift=0}
        },
        zoomboxes below/.code={
            \pgfkeys{/tikz/zoomboxes yshift=-1},
            \pgfkeys{/tikz/zoomboxes xshift=0}
        },
        caption margin/.initial=4ex,
    },
    adjust caption spacing/.code={},
    image container/.style={
        inner sep=0pt,
        at=(image.north),
        anchor=north,
        adjust caption spacing
    },
    zoomboxes container/.style={
        inner sep=0pt,
        at=(image.north),
        anchor=north,
        name=zoomboxes container,
        xshift=\pgfkeysvalueof{/tikz/zoomboxes xshift}*(\imagewidth+\pgfkeysvalueof{/tikz/spymargin}),
        yshift=\pgfkeysvalueof{/tikz/zoomboxes yshift}*(\imageheight+\pgfkeysvalueof{/tikz/spymargin}+\pgfkeysvalueof{/tikz/caption margin}),
        adjust caption spacing
    },
    calculate dimensions/.code={
        \pgfpointdiff{\pgfpointanchor{image}{south west} }{\pgfpointanchor{image}{north east} }
        \pgfgetlastxy{\imagewidth}{\imageheight}
        \global\let\imagewidth=\imagewidth
        \global\let\imageheight=\imageheight
        \gdef\columncount{1}
        \gdef\rowcount{1}
        
    },
    image node/.style={
        inner sep=0pt,
        name=image,
        anchor=south west,
        append after command={
            [calculate dimensions]
            node [image container,subfigurename=\pgfkeysvalueof{/tikz/figurename}-image] {\phantomimage}
            node [zoomboxes container,subfigurename=\pgfkeysvalueof{/tikz/figurename}-zoom] {\phantomimage}
        }
    },
    color code/.style={
        zoombox paths/.append style={draw=#1}
    },
    connect zoomboxes/.style={
    spy connection path={\draw[draw=none,zoombox paths] (tikzspyonnode) -- (tikzspyinnode);}
    },
    help grid code/.code={
        \begin{scope}[
                x={(image.south east)},
                y={(image.north west)},
                font=\footnotesize,
                help lines,
                overlay
            ]
            \foreach \x in {0,1,...,9} { 
                \draw(\x/10,0) -- (\x/10,1);
                \node [anchor=north] at (\x/10,0) {0.\x};
            }
            \foreach \y in {0,1,...,9} {
                \draw(0,\y/10) -- (1,\y/10);                        \node [anchor=east] at (0,\y/10) {0.\y};
            }
        \end{scope}    
    },
    help grid/.style={
        append after command={
            [help grid code]
        }
    },
}

\newcommand\phantomimage{%
    \phantom{%
        \rule{\imagewidth}{\imageheight}%
    }%
}
\newcommand\zoombox[2][]{
    \begin{scope}[zoombox paths]
        \pgfmathsetmacro\xpos{
            (\columncount-1)*(\imagewidth / \pgfkeysvalueof{/tikz/zoomboxarray columns} + \pgfkeysvalueof{/tikz/zoomboxarray inner gap} / \pgfkeysvalueof{/tikz/zoomboxarray columns} ) + \pgflinewidth
        }
        \pgfmathsetmacro\ypos{
            (\rowcount-1)*( \imageheight / \pgfkeysvalueof{/tikz/zoomboxarray rows} + \pgfkeysvalueof{/tikz/zoomboxarray inner gap} / \pgfkeysvalueof{/tikz/zoomboxarray rows} ) + 0.5*\pgflinewidth
        }
        \edef\dospy{\noexpand\spy [
            #1,
            zoombox paths/.append style={
                black and white pattern=\patternnumber
            },
            every spy on node/.append style={#1},
            x=\imagewidth,
            y=\imageheight
        ] on (#2) in node [anchor=north west] at ($(zoomboxes container.north west)+(\xpos pt,-\ypos pt)$);}
        \dospy
        \pgfmathtruncatemacro\pgfmathresult{ifthenelse(\columncount==\pgfkeysvalueof{/tikz/zoomboxarray columns},\rowcount+1,\rowcount)}
        \global\let\rowcount=\pgfmathresult
        \pgfmathtruncatemacro\pgfmathresult{ifthenelse(\columncount==\pgfkeysvalueof{/tikz/zoomboxarray columns},1,\columncount+1)}
        \global\let\columncount=\pgfmathresult
        \ifblackandwhitecycle
            \pgfmathtruncatemacro{\newpatternnumber}{\patternnumber+1}
            \global\edef\patternnumber{\newpatternnumber}
        \fi
    \end{scope}
}

\def\BibTeX{{\rm B\kern-.05em{\sc i\kern-.025em b}\kern-.08em
    T\kern-.1667em\lower.7ex\hbox{E}\kern-.125emX}}
\begin{document}

\title{Exploring Metric Fusion for Evaluation of NeRFs
}

\author{\IEEEauthorblockN{Shreyas Shivakumara}
\IEEEauthorblockA{\textit{Department of Science and Technology } \\
\textit{Linköping University}\\
Norrköping, Sweden \\
shreyas.shivakumara@liu.se}
\and
\IEEEauthorblockN{Gabriel Eilertsen}
\IEEEauthorblockA{\textit{Department of Science and Technology } \\
\textit{Linköping University}\\
Norrköping, Sweden \\
gabriel.eilertsen@liu.se}
\and
\IEEEauthorblockN{Karljohan Lundin Palmerius}
\IEEEauthorblockA{\textit{Department of Science and Technology } \\
\textit{Linköping University}\\
Norrköping, Sweden \\
karljohan.lundin.palmerius@liu.se}
}

\maketitle

\begin{abstract}

Neural Radiance Fields (NeRFs) have demonstrated significant potential in synthesizing novel viewpoints.  
Evaluating the NeRF-generated outputs, however, remains a challenge due to the unique artifacts they exhibit, and no individual metric performs well across all datasets. We hypothesize that combining two successful metrics, Deep Image Structure and Texture Similarity (DISTS) and Video Multi-Method Assessment Fusion (VMAF), based on different perceptual methods, can overcome the limitations of individual metrics and achieve improved correlation with subjective quality scores. We experiment with two normalization strategies for the individual metrics and two fusion strategies to evaluate their impact on the resulting correlation with the subjective scores. The proposed pipeline is tested on two distinct datasets, Synthetic and Outdoor, and its performance is evaluated across three different configurations. We present a detailed analysis comparing the correlation coefficients of fusion methods and individual scores with subjective scores to demonstrate the robustness and generalizability of the fusion metrics. 

 \end{abstract}


\section{Introduction}

The Neural Radiance Field (NeRF) is a 3D reconstruction technique that models a volumetric function to synthesize novel views of a scene from a sparse set of images \cite{mildenhall2020nerf}. It has gained significant attention due to its ability to render scenes with high fidelity and robustness. Applications of NeRF span from large-scale 3D reconstruction of urban environments to the creation of immersive content for augmented and virtual reality applications \cite{blockNerf}\cite{xu2023vr}. 

\begin{figure}
\centering

\begin{tikzpicture}[
    zoomboxarray,
    zoomboxes right,
    zoombox paths/.style={draw=blue, thick},
    every spy on node/.style={zoombox paths},
    every spy in node/.style={zoombox paths}
]
    \node [image node] {\includegraphics[width=0.14\textwidth]{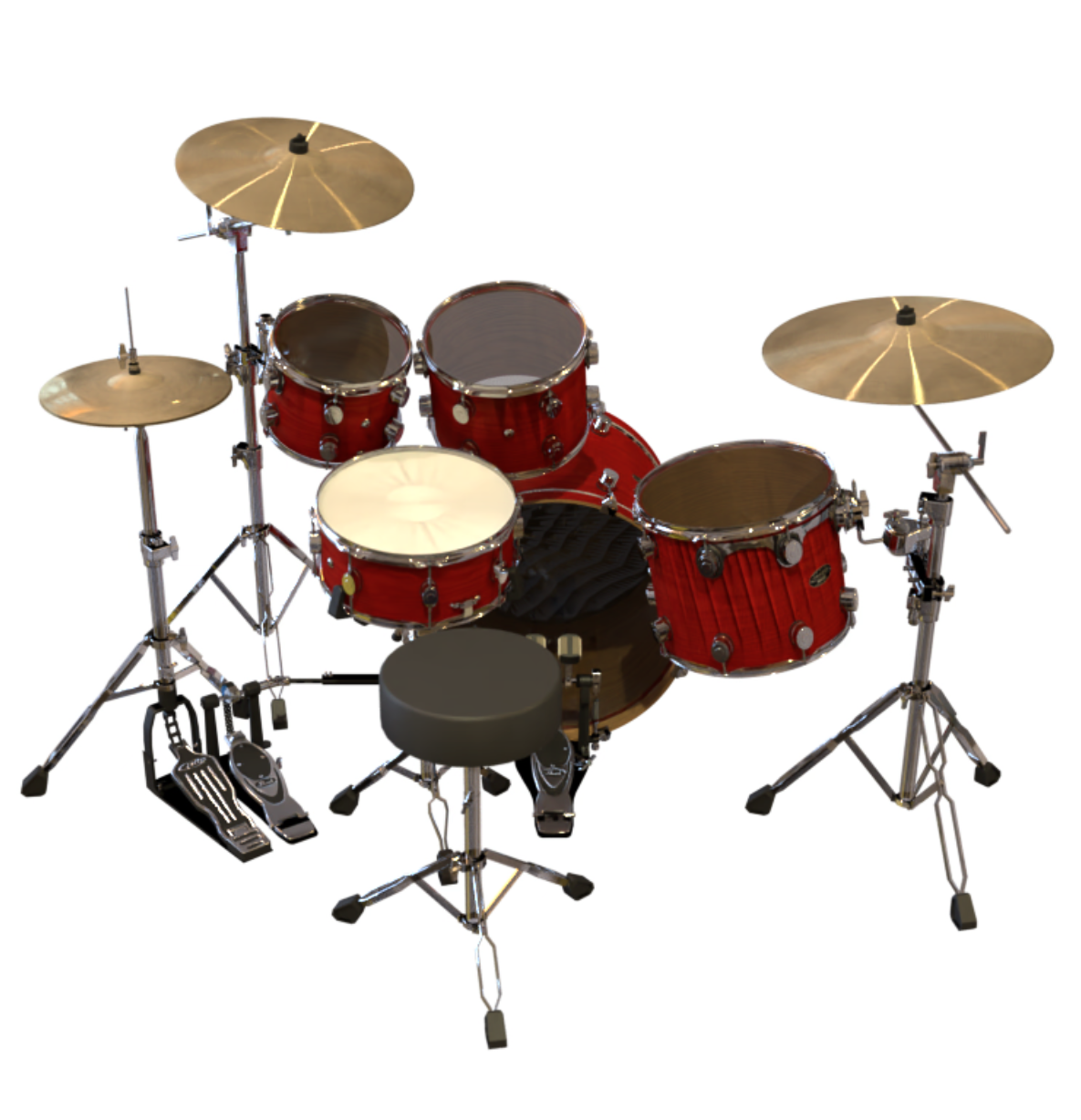}};
    \zoombox[magnification=3]{0.27,0.83}
    \zoombox[magnification=3]{0.7,0.48}
    \zoombox[magnification=3]{0.38,0.55}
    \zoombox[magnification=3]{0.46,0.7}
\end{tikzpicture}
\vspace{1em}
\begin{tikzpicture}[
    zoomboxarray,
    zoomboxes right,
    zoombox paths/.style={draw=red, thick},
    every spy on node/.style={zoombox paths},
    every spy in node/.style={zoombox paths}
]
    \node [image node] {\includegraphics[width=0.14\textwidth]{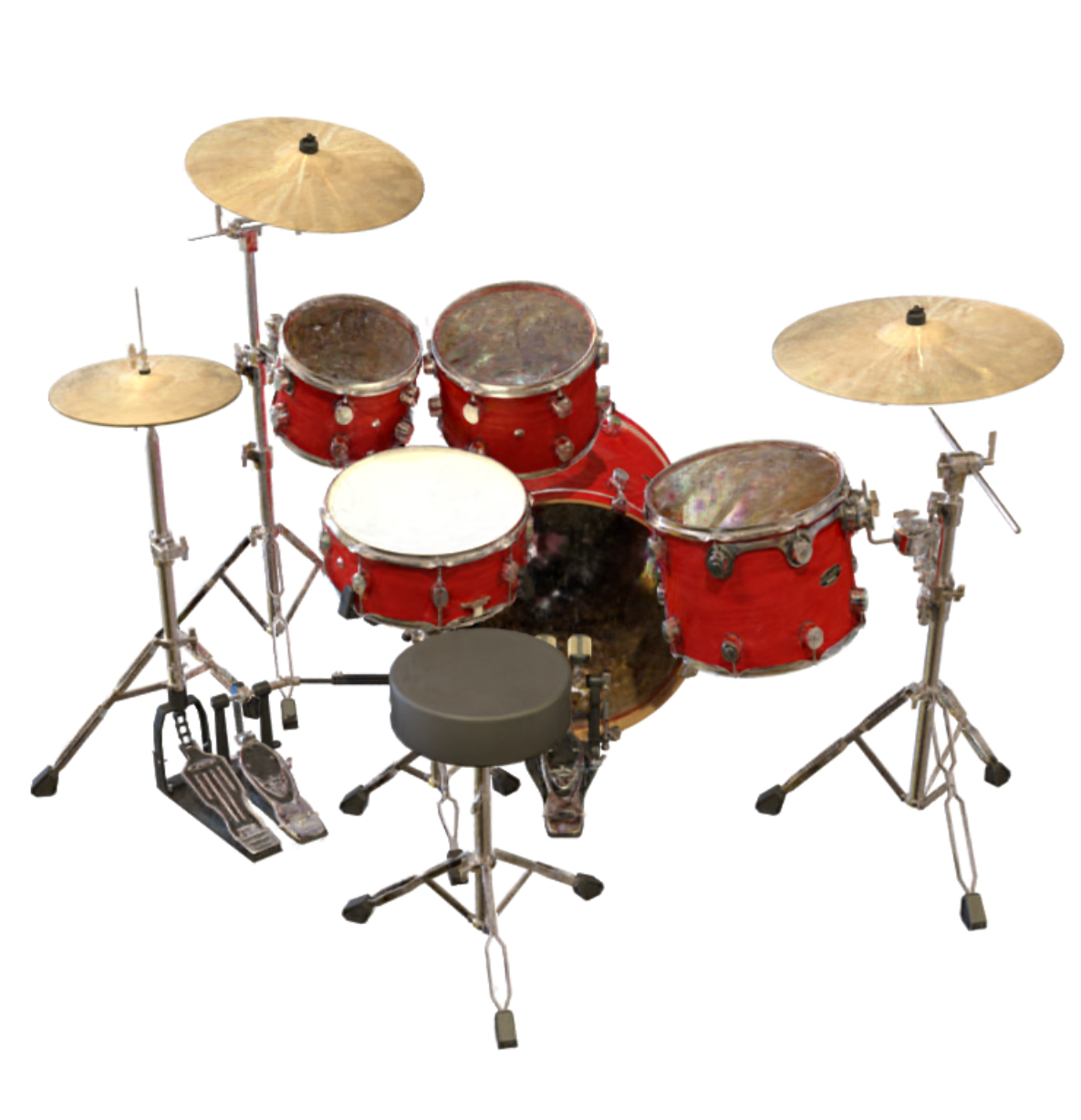}};
    \zoombox[magnification=3]{0.27,0.83}
    \zoombox[magnification=3]{0.7,0.48}
    \zoombox[magnification=3]{0.38,0.55}
    \zoombox[magnification=3]{0.46,0.7}
\end{tikzpicture}

\caption{Reference (top) and synthesized (below) images of a drum scene, with zoomed-in patches highlighting visual differences. VMAF (82.0/100; higher is better) and DISTS (0.12/1; lower is better) shows an example where VMAF fails and DISTS captures the textural artifacts.}
\label{zoom}
\end{figure}

 EVALUAT The evaluation of NeRF-synthesized outputs is performed using metrics such as Peak Signal-to-Noise Ratio (PSNR), Structural Similarity Index Measure (SSIM), and Learned Perceptual Image Patch Similarity (LPIPS), largely due to their ease of implementation and widespread adoption. However, studies have shown that these metrics fail to detect artifacts and show poor correlation with subjective scores for NeRF-synthesized outputs \cite{liang2023}. In contrast, studies have demonstrated that video-based metrics like VMAF and learning-based metrics such as DISTS have consistently shown high correlations with subjective evaluations \cite{explicit_nerf_qa_2024}\cite{martin2024nerfviewsynthesissubjective}\cite{onuoha2023}.  Nonetheless, no single metric currently exists that consistently correlates fully with subjective scores across diverse datasets as illustrated by an example in Figure \ref{zoom}. This gap motivates the exploration of fusion methods that combine complementary strengths of two metrics to evaluate NeRF-synthesized outputs. While similar work on aggregating metrics for image quality assessment has been investigated in \cite{multi_metric}, no such approaches have been applied to NeRFs to date.


In this study, we investigate the fusion of two successful complementary metrics in evaluating NeRF-synthesized outputs, DISTS and VMAF. To ensure comparability between metrics, we first apply two normalization techniques: min-max scaling and z-score standardization. We then explore merging using two fusion strategies: minimum selection and averaging. We hypothesize that merging using the right normalization technique will provide an improved correlation with subjective scores compared to using individual metrics alone. We test the pipeline on two distinct datasets under three different configurations to evaluate the robustness and generalizability.

\section{Datasets}

The two datasets used for evaluating the pipeline are chosen for their diversity in camera orientations, complexity and availability of subjective quality scores. 

The first dataset is from \cite{explicit_nerf_qa_2024}, named \emph{Synthetic}(S), which includes 22 synthetic scenes such as Lego, Drums, and Photograph. The dataset contains 440 videos synthesized from NeRF models such as \cite{instantngp} and \cite{plenoxels} across different compression configurations. The subjective experiment used a double stimulus impairment scale with 21 participants and reported scores using the Mean Opinion Score (MOS). Due to missing MOS values in two scenes, we selected 20 scenes (400 videos) for the analysis.  

The second dataset is from \cite{martin2024nerfviewsynthesissubjective}, including a mix of synthetic indoor and 16 real outdoor scenes. Since the indoor scenes overlap with those in synthetic dataset, we added only the outdoor scenes in our analysis, which we refer to as \emph{Outdoor} (O). Example scenes include synthesized reconstructions such as Train and Truck, generated using NeRF models such as \cite{barren_mipnerf3602022} and \cite{nerfstudio}. The subjective evaluations were performed using the double stimulus continuous quality scale methodology, reported with Difference Mean Opinion Scores (DMOS), collected from 22 participants (16 male and 6 female). To ensure consistency of subjective scores with the first dataset, we approximated MOS scores from the DMOS values\cite{itu_2023}.

\section{Metric Fusion Pipeline}

\begin{figure}[htbp]
\centerline{\includegraphics [width=0.9\linewidth]{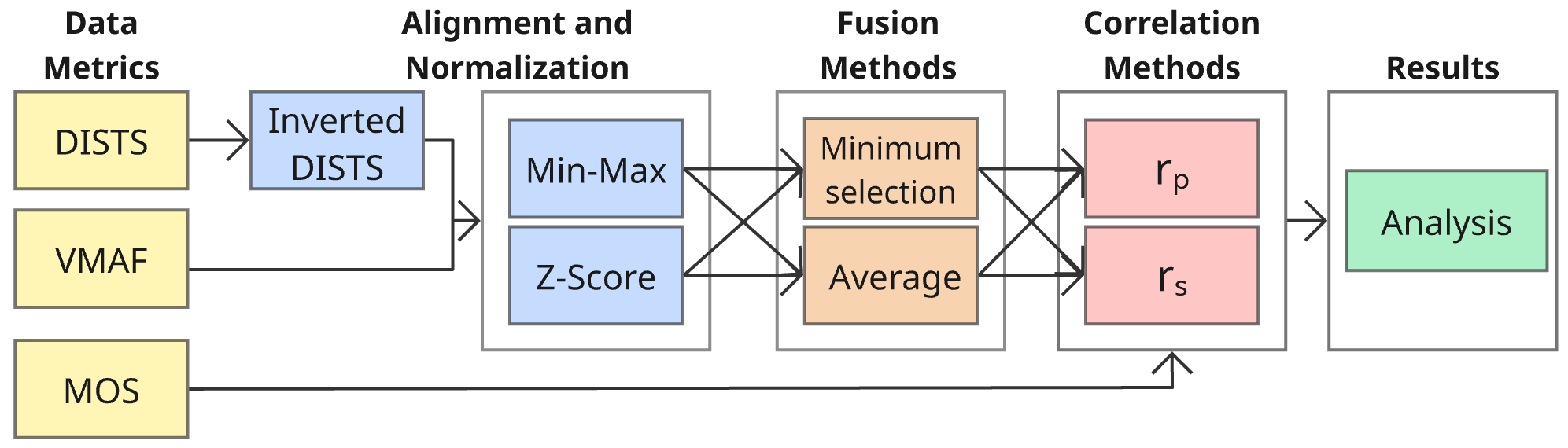}}
\caption{Metric fusion pipeline}
\label{pipeline}
\label{fig}
\end{figure}

The core components of the pipeline for metric fusion are illustrated in Figure \ref{pipeline}.

We are primarily focusing on two metrics: DISTS \cite{dists2020} and VMAF\cite{vmaf2016}, due to their consistent high correlation with subjective scores in NeRF synthesis, as demonstrated in \cite{explicit_nerf_qa_2024},\cite{martin2024nerfviewsynthesissubjective} and \cite{onuoha2023}. DISTS is a full-reference, learning-based image quality metric known for its robustness to texture variation and geometric distortions\cite{dists2020}. VMAF, on the other hand, is a full-reference video quality assessment metric that captures temporal coherence to better align with human perception of video quality\cite{vmaf2016}. 
 
We have inverted the DISTS scores so the higher values indicate better quality, aligning with MOS and VMAF. Since the two metrics are in different ranges, a normalization step is necessary. We use two normalization approaches, min-max (mm) and z-score (z) in our experiment.

Min-Max normalization is a linear transformation applied to scale the data $x$ to the range [0,1] and is formulated as: \[ x' = \frac{x - \min(x)}{\max(x) - \min(x)}, \] where $\min(x)$ and $\max(x)$ are the minimum and maximum values of the data, respectively, and $x'$ is the normalized value. 

Z-score is a standardization techniques that transforms the data to have zero mean and unit standard deviation, denoted by:
\[
z = \frac{x - \mu}{\sigma},
\]
where \( \mu \) is the mean of the dataset, \( \sigma \) is the standard deviation of the dataset,
and \( z \) represents the normalized value.


We perform two distinct fusion strategies over VMAF and DISTS. The first fusion strategy is the minimum selection (min), which chooses the minimum value of the two metrics. We consider that the human perception is sensitive to artifacts and thus, the overall quality scores are dominated by the minimum metric value. The second fusion strategy is the average (avg), considering equal contribution from both metrics. This approach allows us to evaluate how the fusion scores behave when VMAF and DISTS are given equal emphasis.





\section{Evaluation and Results}

\begin{table}[htbp]
\caption{Summary of Fusion Correlation Results}
\label{merged_results}
\centering
\begin{tabular}{|l|c|l|c|c|}
\hline
\textbf{Dataset} & \textbf{\#Samples(C/T)} & \textbf{Fusion} & \textbf{$r_p$ mean} & \textbf{$r_s$ mean} \\
\hline
Calib - S & 320 / 80 & avg\_mm & 0.837 & 0.826 \\
Test - S & & min\_mm & 0.720 & 0.704 \\
 & & avg\_z & \textbf{0.883} & \textbf{0.867} \\
 & & min\_z & 0.800 & 0.808 \\
\hline
Calib - O & 12 / 4 & avg\_mm & 0.990 & 1.000 \\
Test - O & & min\_mm & 0.994 & 1.000 \\
 & & avg\_z & 0.991 & 1.000 \\
 & & min\_z & 0.992 & 1.000 \\
\hline
Calib - S & 400 / 16 & avg\_mm & \textbf{0.788} & \textbf{0.797} \\
Test - O & & min\_mm & 0.770 & 0.716 \\
 & & avg\_z & 0.734 & 0.686 \\
 & & min\_z & 0.770 & 0.716 \\
\hline
\end{tabular}
\end{table}

To ensure a fair and consistent comparison, the dataset is split into a calibration set and a test set by unique scenes. Normalization scalers are fitted on the calibration dataset and applied to the test dataset to align metric scales before fusion. We evaluate the method in three configurations. The first configuration uses a synthetic dataset, splitting the data into 80\% calibration and 20\% test. The second configuration applies the same 80/20 split on the outdoor dataset. The third configuration is a test to evaluate cross-validation and generalizability. The scalers are calibrated on the whole synthetic dataset and tested on the whole outdoor dataset. The normalized metrics are then combined using the two fusion strategies. For each sample, we calculate the Pearson correlation coefficient ($r_p$) \cite{pearson} and Spearman rank correlation coefficient ($r_s$) \cite{spearman} to measure the relationship between metric scores (both individual and fused) and subjective scores. Since we are comparing quality metrics, we perform bootstrap sampling to estimate the distribution and variance of the correlation values \cite{bootstrapping}. For each run, we generate 1000 bootstrap samples by randomly sampling scenes with replacements from the test dataset. We then use the distributions of $r_p$ and $r_s$ to compute the mean and 95\% confidence intervals, which quantify the uncertainty in the correlation estimates.
 
\begin{figure}[htbp]
\centering

\begin{subfigure}[b]{\linewidth}
    \centering
    \includegraphics[width=\linewidth]{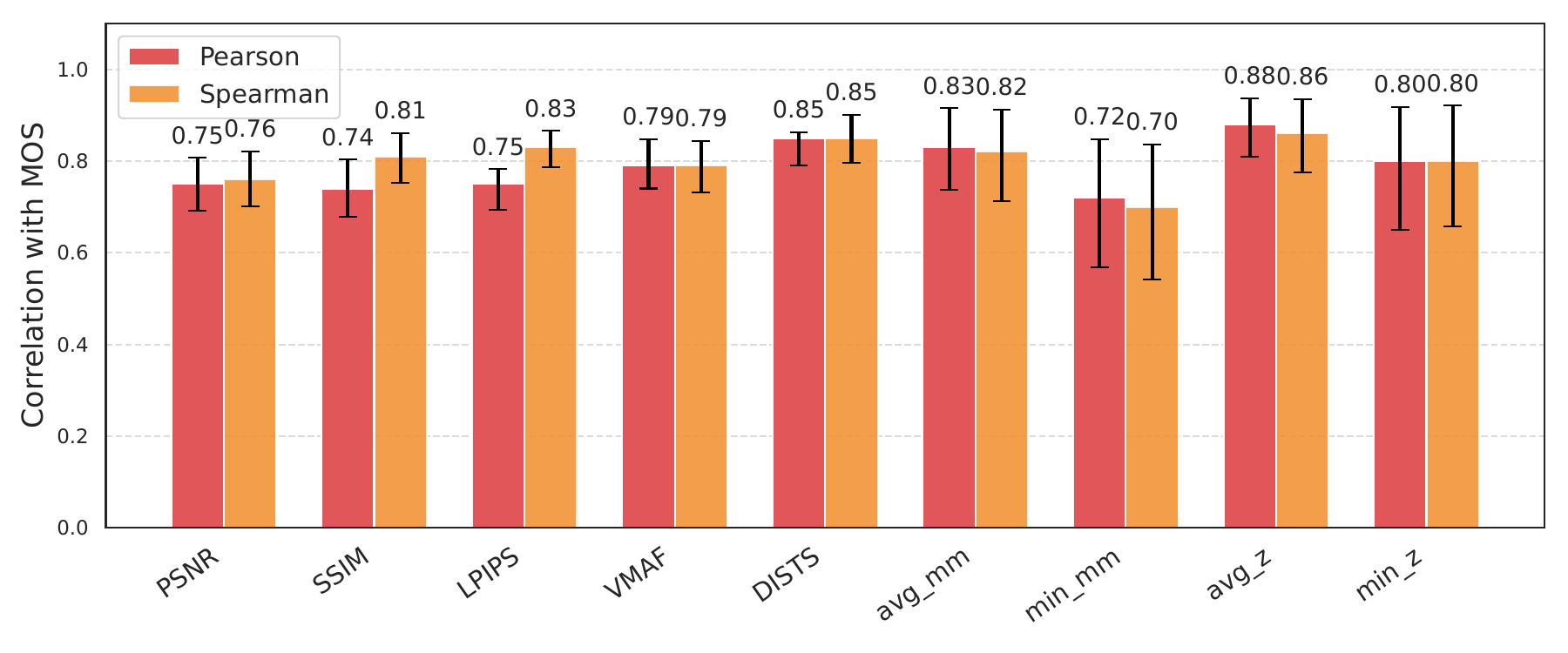}
    \caption{Configuration 1 - Calibrated and tested on Synthetic dataset}
    \label{synthetic_correlation_graph}
\end{subfigure}

\vspace{0.5em}  

\begin{subfigure}[b]{\linewidth}
    \centering
    \includegraphics[width=\linewidth]{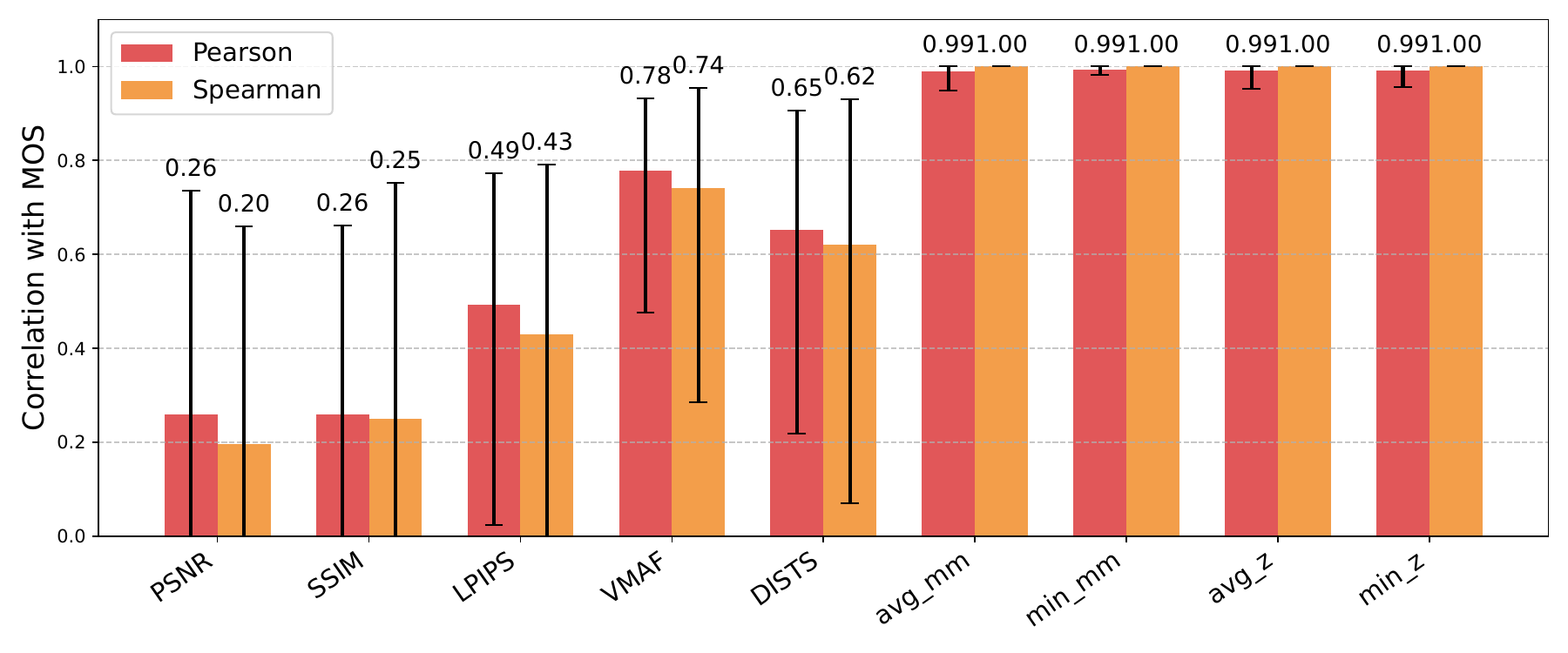}
    \caption{Configuration 2 - Calibrated and tested on Outdoor dataset}
    \label{new_outdoor_correlation_graph}
\end{subfigure}

\vspace{0.5em}  

\begin{subfigure}[b]{\linewidth}
    \centering
    \includegraphics[width=\linewidth]{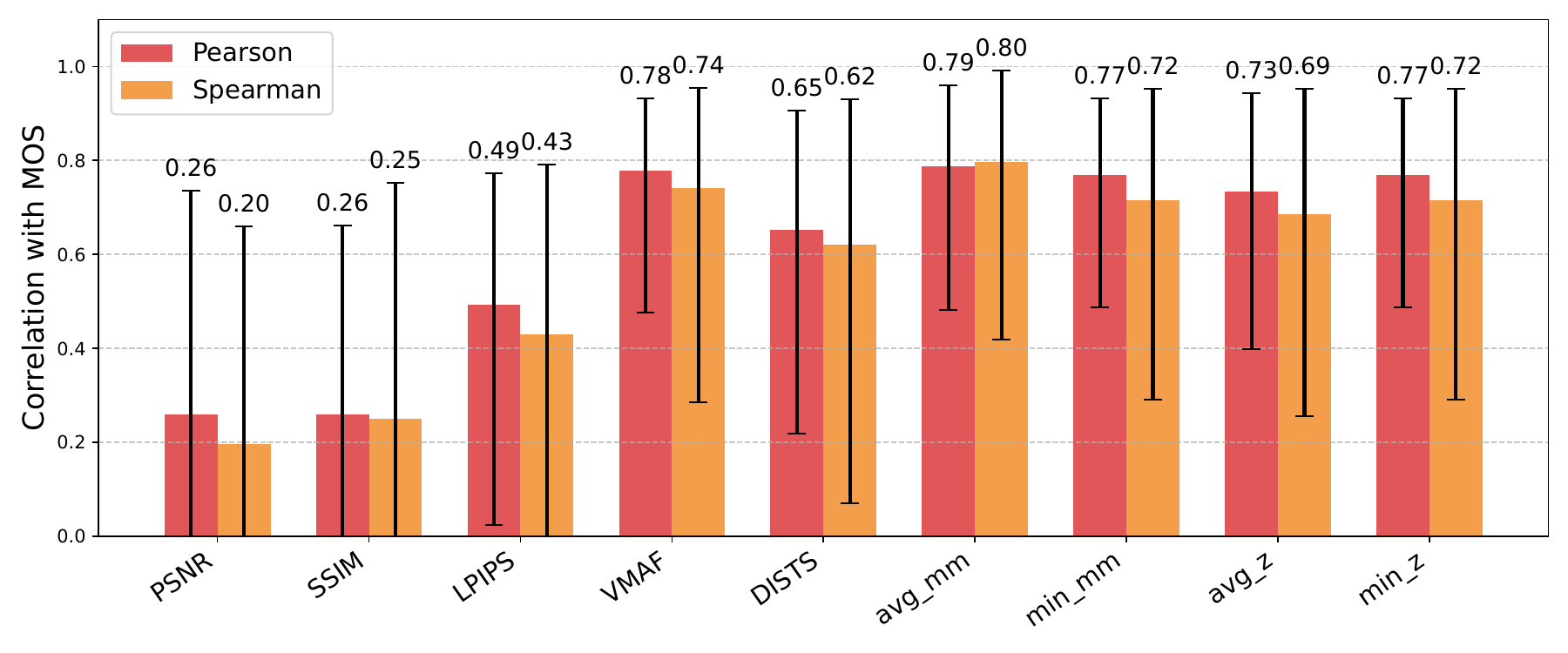}
    \caption{Configuration 3 - Calibrated on Synthetic dataset and Tested on Outdoor dataset}
    \label{outdoor_correlation_graph}
\end{subfigure}

\caption{
Mean Pearson and Spearman correlation coefficients of different configurations, comparing fusion strategies and individual metrics. Fusion scores are named as fusion\_method\_normalization\_technique. Error bars indicate 95\% confidence intervals.
}
\label{fig:all_configurations_correlation}
\end{figure}

The evaluation results across three different configurations is presented in Table \ref{merged_results}. Figures \ref{fig:all_configurations_correlation} shows the correlation scores along with estimated confidence intervals for all the configurations. We have also considered the most common metrics such as PSNR, SSIM and LPIPS in evaluating NeRF methods to highlight the difference between the individual metrics and fusion methods of VMAF and DISTS.

\section{Discussion}

From the results of Configuration 1, avg\_mm  and avg\_z fusion methods show a slight improvement compared to DISTS, in correlation with subjective scores, but these enhancements are marginal compared to those achieved by individual metrics. This indicates that the individual metrics perform sufficiently well on datasets with lower complexity, effectively detecting perceptual artifacts. Conversely, in the second configuration, both Pearson and Spearman correlations are very strong. However, these high correlations may be caused by the relatively small sample size and potential overfitting of the metric calibration to the dataset. Therefore, it may not be appropriate to draw definitive conclusions from this configuration alone. Results from the third configuration, evaluating cross-validation, the mean correlations show that avg\_mm outperforms individual metrics and other fusion methods. The mean performance of the fusion methods shows improvement but the considerable variance indicates that they may fail to capture individual cases that exhibit significant fluctuations. This indicates that scalers calibrated on a simpler dataset and tested on a more complex dataset improve mean correlation with subjective scores, demonstrating the robustness and generalizability of the fusion.

\begin{figure}[htbp]
  \centering
  \begin{minipage}{0.48\linewidth}
    \centering
    \includegraphics[width=\linewidth]{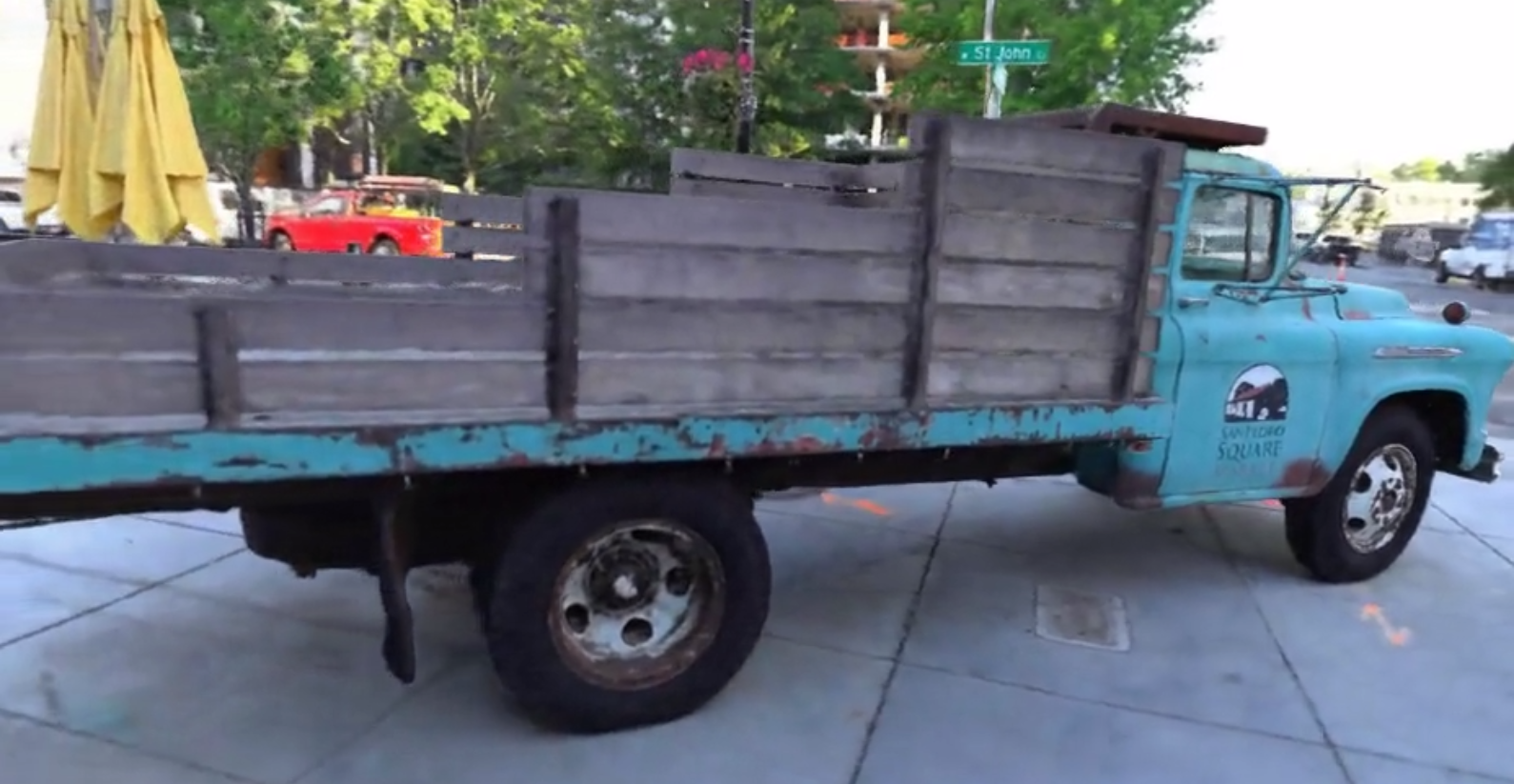}
    \label{fig:first}
  \end{minipage}
  \hfill
  \begin{minipage}{0.48\linewidth}
    \centering
    \includegraphics[width=\linewidth]{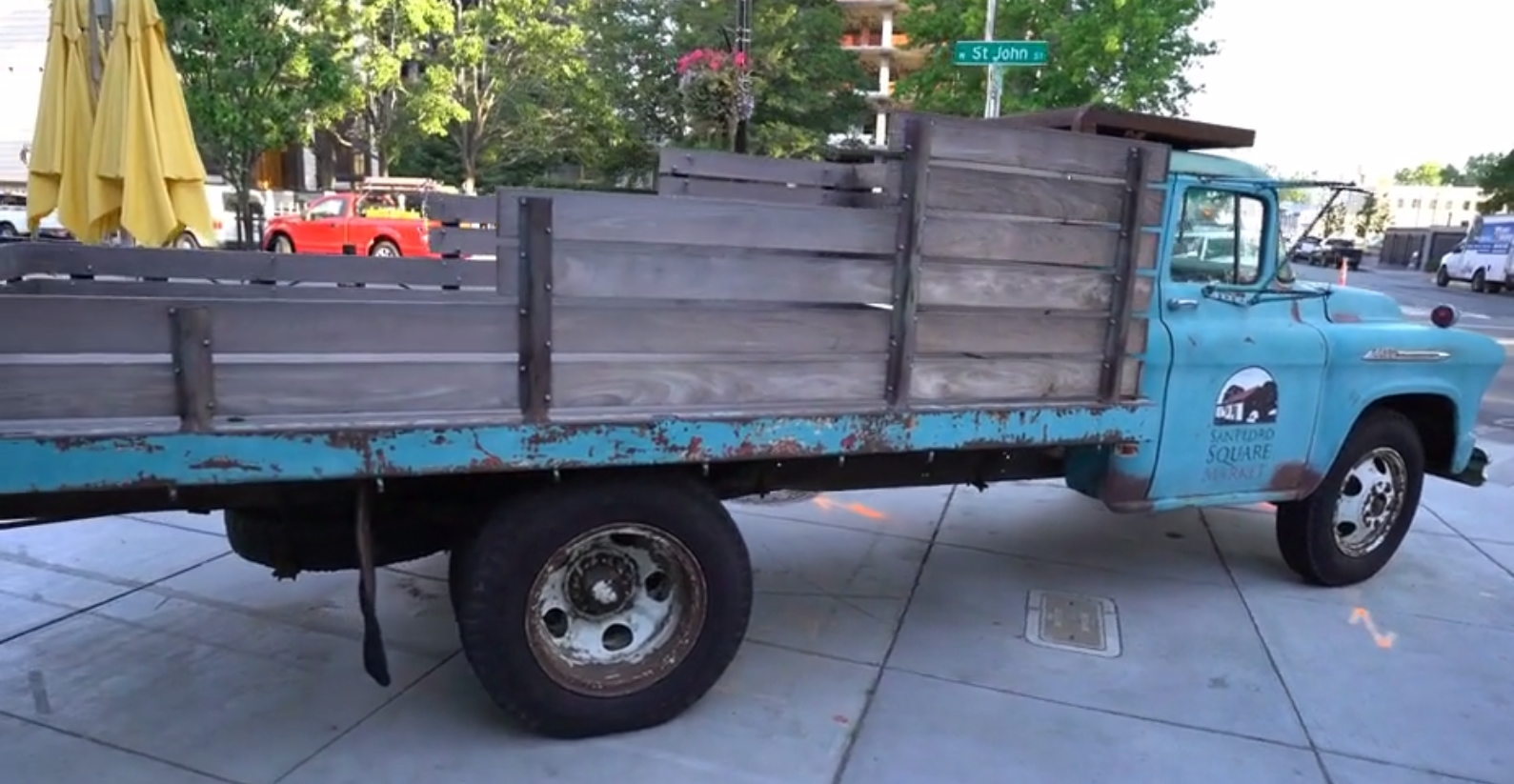}  
    \label{fig:second}
  \end{minipage}
  \caption{Synthesized (left) and reference (right) images of an outdoor truck scene. VMAF (13.0/100) detects artifacts missed by DISTS (0.097/1), these artifacts are sensitive to VMAF metric but negligible to humans and the fusion method avg\_mm (0.8) correlates better with subjective scores.}
\label{example_truck}
\end{figure}

Performance variations across normalization techniques and fusion methods indicate that avg\_mm consistently correlated better to subjective scores. avg\_mm, given the equal emphasis that reflects the complementary strengths of VMAF, which captures temporal artifacts, and DISTS, which detects diverse spatial perceptual distortions, can identify and correlate better with subjective scores. Figure \ref{example_truck} illustrates an example where avg\_mm performs better compared to individual metrics on a truck scene.   

\section{Conclusion}

Evaluating the perceptual quality of NeRF-synthesized output remains challenging, as no single metric performs consistently across all datasets. We experiment merging two successful complementary metrics, VMAF and DISTS, using minimum selection and averaging fusion strategies and explore different normalization techniques to ensure the metrics are comparable. We evaluate the methods on three different configurations and our results demonstrate that combining these metrics with appropriate normalization may provide a robust and reliable approach compared to an individual metric for NeRF quality assessment across diverse datasets. Future work will explore incorporating additional metrics and advanced fusion models, such as ridge regression or neural networks, to learn optimal fusion weights for better alignment with subjective scores.



\bibliographystyle{IEEEtran}
\bibliography{main}
\end{document}